\documentclass[journal]{IEEEtran}
\usepackage{amsmath,amsfonts}
\usepackage{algorithmic}
\usepackage{array}
\usepackage[caption=false,font=normalsize,labelfont=sf,textfont=sf]{subfig}
\usepackage{textcomp}
\usepackage{stfloats}
\usepackage{url}
\usepackage{verbatim}
\usepackage{graphicx}
\def\BibTeX{{\rm B\kern-.05em{\sc i\kern-.025em b}\kern-.08em
    T\kern-.1667em\lower.7ex\hbox{E}\kern-.125emX}}
\usepackage{balance}

\usepackage{cite}
\usepackage{cleveref}
\usepackage{tipa}
\usepackage{booktabs}
\usepackage{multirow}
\usepackage{adjustbox}
\usepackage{listings}
\usepackage{placeins}
\usepackage{colortbl,xcolor}

\definecolor{drmcolor}{HTML}{FFA500}
\definecolor{yeocolor}{HTML}{000080}
\definecolor{colorint}{RGB}{100, 143, 255}

\begin{document}
\title{Multilingual Dysarthric Speech Assessment Using Universal Phone Recognition and Language-Specific Phonemic Contrast Modeling}
\author{Eunjung Yeo, 
Julie M. Liss, 
Visar Berisha, 
David R. Mortensen 
\thanks{Eunjung Yeo is with the Department of Computer Science, University of Texas at Austin, Austin, TX, USA (email: eunjung.yeo@utexas.edu).
Julie M. Liss is with the College of Health Solutions, Arizona State University, Tempe, AZ, USA.
Visar Berisha is with the School of Electrical, Computer and Energy Engineering and the College of Health Solutions, Arizona State University, Tempe, AZ, USA.
David R. Mortensen is with the Language Technologies Institute, School of Computer Science, Carnegie Mellon University, Pittsburgh, PA, USA.}
}

\markboth{IEEE Transactions on Audio, Speech and Language Processing}%
{How to Use the IEEEtran \LaTeX \ Templates}

\maketitle

\begin{abstract}
The growing prevalence of neurological disorders associated with dysarthria motivates the need for automated intelligibility assessment methods that are applicalbe across languages. However, most existing approaches are either limited to a single language or fail to capture language-specific factors shaping intelligibility. We present a multilingual phoneme-production assessment framework that integrates universal phone recognition with language-specific phoneme interpretation using contrastive phonological feature distances for phone-to-phoneme mapping and sequence alignment. The framework yields three metrics: phoneme error rate (PER), phonological feature error rate (PFER), and a newly proposed alignment-free measure, phoneme coverage (PhonCov). 
Analysis on English, Spanish, Italian, and Tamil show that PER benefits from the combination of mapping and alignment, PFER from alignment alone, and PhonCov from mapping. Further analyses demonstrate that the proposed framework captures clinically meaningful patterns of intelligibility degradation consistent with established observations of dysarthric speech.

\end{abstract}


\section{Introduction}
Neurological disorders such as Parkinson’s disease (PD), amyotrophic lateral sclerosis (ALS), and stroke are commonly associated with dysarthria, a motor speech disorder resulting from impaired neuromotor control of the speech production mechanism. Dysarthria disrupts the coordination, timing, and precision of articulatory movements, often accompanied by abnormal prosody and reduced speech intelligibility, which substantially impacts functional communication \cite{darley1969differential}. Consequently, reliable and objective intelligibility assessment is essential for clinical diagnosis, treatment planning, and longitudinal monitoring of progression and therapeutic outcomes.

As the global prevalence of neurological disorders associated with dysarthria increases \cite{WHO2024}, there is a growing need for assessment tools that can be easily scaled across languages. However, current clinical practice relies on perceptual ratings from speech-language pathologists (SLPs), which are subjective \cite{hirsch2022reliability}, time-intensive, and difficult to scale.
Furthermore, many established assessment protocols are primarily grounded in English-centric assumptions, limiting cross-linguistic applicability to other languages  \cite{liss2013crosslinguistic, kim2017cross, kim2024introduction}.

\begin{figure}[t]
    \centering
    \includegraphics[width=0.9\columnwidth]{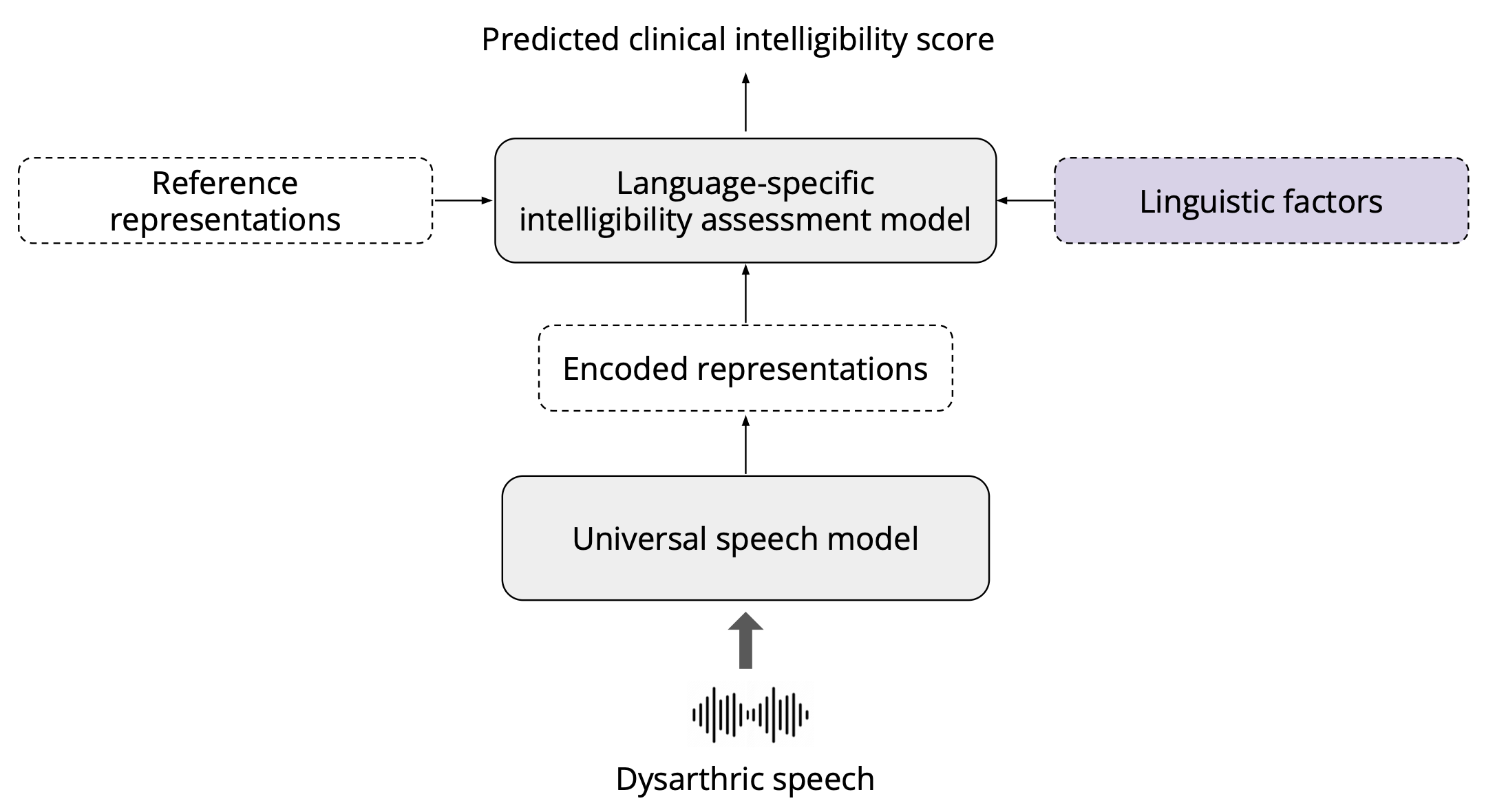}
    \caption{Conceptual framework for multilingual dysarthric speech assessment proposed in \cite{yeo2025applications}. The universal speech model encodes language-universal speech manifestations, while the language-specific intelligibility assessment model interprets the encoded representations based on the structure of each language. 
    }  
    \label{fig:fig0}
\end{figure}
Research in automatic intelligibility assessment aims to overcome the limits of perceptual evaluation by predicting intelligibility scores that correlate well with SLP ratings. However, most existing approaches remain monolingual, reflecting practical constraints in the availability of speech data (both healthy and dysarthric), clinical annotations, and the computational resources required for model training and maintenance. 

To address these limitations, researchers have explored multilingual models that leverage language-universal characteristics \cite{kovac2022exploring, favaro2023multilingual, veetil2024robust}. Nevertheless, recent studies have shown that universal representations alone are insufficient: multilingual systems frequently underperform their monolingual counterparts \cite{kovac2022exploring, yeo2022cross}, even when trained on larger and more diverse datasets (details in \Cref{ssec:related-multilingual}).

To reconcile scalability with linguistic specificity, a principled solution was proposed in \cite{yeo2025applications}. The proposed two-stage framework first encodes speech into a language-universal representation and then applies language-specific modules that interpret this representation based on linguistic factors (\Cref{fig:fig0}), such as phonemic contrasts, rhythmic typologies, and tonal distinctions. This design enables multilingual assessment without requiring the training of separate models for each language, while preserving sensitivity to the linguistic structure that shapes intelligibility within each language.

The present work instantiates this framework at the phoneme-production level\footnote{Although dysarthria intelligibility is influenced by both segmental and suprasegmental factors and the framework in \cite{yeo2025applications} addresses both, we focus on phoneme-level production as the first instantiation of the framework.} and empirically examines its impact on intelligibility assessment of dysarthric speech.
The proposed approach uses a Universal Phone Recognizer (UPR) to obtain language-universal phonetic transcriptions, which are then interpreted with respect to each language’s phoneme inventory using linguistically grounded, contrast-aware phoneme mapping and alignment. This processing supports the computation of clinically interpretable phoneme-level metrics, including phoneme error rate (PER), phoneme feature error rate (PFER), and phoneme coverage (PhonCov), which are evaluated against expert SLP ratings.

This work makes the following contributions:
\begin{itemize}
\item We present a multilingual phoneme-production assessment pipeline that integrates universal phone recognition with language-specific phoneme interpretation for dysarthric speech.
\item We propose \textit{PhonCov}, an alignment-free phoneme-level metric that quantifies reductions in a speaker’s phoneme inventory, complementing PER and PFER.
\item We empirically evaluate the impact of language-specific phoneme interpretation on phoneme-production metrics across four languages, demonstrating improvements in their association with SLP-rated intelligibility.
\end{itemize}

\section{Related Work}\label{sec:relatedwork}
\subsection{Automatic Dysarthria Assessment} \label{ssec:background}
Research on automatic dysarthria assessment has sought to balance predictive performance with clinical interpretability, leading to two approaches: hand-crafted acoustic features and neural-based methods \cite{favaro2023interpretable}. Hand-crafted approaches use clinically motivated measures targeting established manifestations of dysarthria, including voice-quality and glottal features \cite{prabhakera2018dysarthric, narendra2019dysarthric}, articulatory and pronunciation-based measures \cite{yeo2021automatic, yeo2023speech, jeyaraman2024pronunciation, thompson2023vowel}, and prosodic cues such as speech rate, pitch, and rhythm \cite{martens2013automated, hernandez2020prosody}. These features may be used individually or combined to improve prediction performance \cite{le2014modeling, favaro2023multilingual}.

Neural approaches model dysarthric speech directly from waveforms or spectrograms. While these approaches have shown promising results \cite{zaidi2021deep, joshy2022automated}, they typically require large amounts of labeled training data, which are difficult to obtain for dysarthric speech. Recent self-supervised learning (SSL) models, combined with fine-tuning, reduce data requirements and demonstrate strong cross-dataset performance \cite{favaro2023interpretable, yeo2023automatic}. However, the representations learned by neural systems are often difficult to interpret and require additional explainability methods to relate model outputs to clinically meaningful speech deficits \cite{montalbo2025dysarnet, xu2025evaluating}.


\subsection{Phoneme Production Assessment}
While dysarthria affects multiple dimensions of speech, including articulation, prosody, and voice quality \cite{darley1969differential}, phoneme pronunciation has played a central role in the assessment of dysarthric speech \cite{kent1989toward, xue2023assessing}. In fact, phonetic transcription using the International Phonetic Alphabet (IPA) is often employed in clinical settings \cite{ball2021transcribing}. 

In clinical speech science, the Percentage of Correct Consonants (PCC) is a widely used as a objective measure of speech intelligibility, quantifying the proportion of correctly pronounced consonants relative to expected number of consonants \cite{shriberg1997percentage, sell2020percent}. In addition to transcription-based measures, acoustic indicators have been used to assess intelligibility in dysarthric speech, such as Vowel Space Area and the F2-ratio, which reflect vowel centralization \cite{kim2011vowel, lansford2014vowel}, as well as Cepstral Peak Prominence (CPP), an acoustic indicator of impaired laryngeal function \cite{maffei2023acoustic, murton2023validation}.

\begin{figure}[t]
    \centering
    \includegraphics[width=\columnwidth]{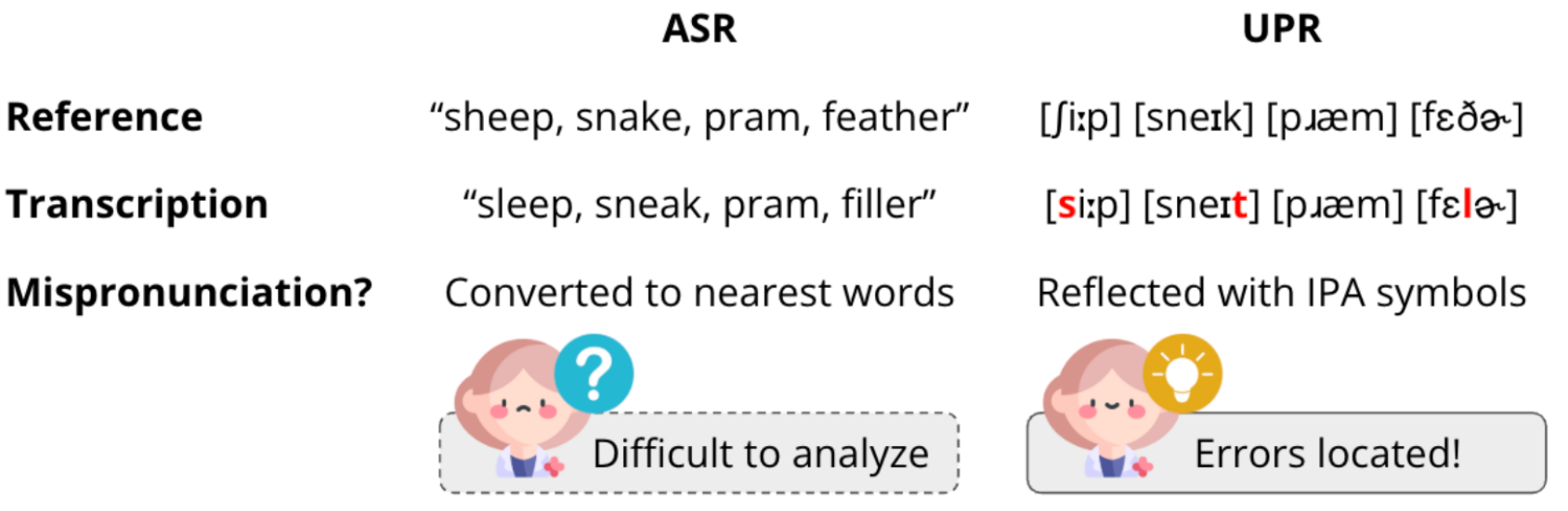}
    \caption{Example of automated pronunciation assessment using ASR and UPR. Symbolic outputs of UPR reveal segmental errors that directly reflect articulatory impairment, such as [\textipa{S}]$\rightarrow$[\textipa{s}] or [\textipa{k}]$\rightarrow$[\textipa{t}].}  
    \label{fig:fig1}
\end{figure}

A conventional computational approach to phoneme production assessment is Goodness of Pronunciation (GoP) \cite{witt2000phone}, which estimates phone-level accuracy using acoustic likelihoods from models trained on healthy speech. However, GoP relies on language-specific acoustic models and phone-level alignment, limiting scalability to multilingual settings. Although SSL models have reduced dependence on language-specific acoustic models \cite{yeo2023speech}, they still require time-aligned phone annotations, which are difficult to obtain for dysarthric speech. Automatic speech recognizers (ASR) have also been used, with Word Error Rate (WER) serving as a proxy for segmental intelligibility \cite{XUE202323}. However, ASR outputs are word-based, limiting their clinical interpretability (\Cref{fig:fig1}).

\subsection{Multilingual Assessment for Dysarthric Speech}\label{ssec:related-multilingual}
Historically, dysarthria has been treated as a language-agnostic disorder, assuming that neuromotor deficits manifest similarly across languages \cite{miller2014motor, kim2024introduction, levy2024revisiting}. However, a growing body of cross-linguistic evidence challenges this assumption, \textit{i.e.} differences in language structure can cause identical neuromotor deficits to impact intelligibility in language-specific ways \cite{liss2013crosslinguistic, kim2017cross, hernandez2020dysarthria, kim2024introduction, yeo2025applications}. 

Cross-linguistic evidence\footnote{There is a terminology difference between the clinical speech science and speech technology communities in the use of the term \textit{cross-lingual}. In clinical sciences, cross-lingual typically refers to comparative analyses across languages, while multilingual refers to individuals who speak multiple languages. In contrast, in speech technology, cross-lingual denotes models trained in one language and tested in another, whereas multilingual refers to models trained on multiple languages and designed to operate on those languages. In this paper, we adopt the speech-technology convention and use the term \textit{multilingual}.} indicates that the perceptual impact of dysarthria is shaped by language-specific phonological and prosodic structures. For example, voice onset time constituted a critical cue for intelligibility assessment in Korean due to its three-way laryngeal contrast, whereas it contributed less to intelligibility judgments in English \cite{kim2017cross}. Likewise, prominence cues differed across languages, with pitch-based prominence playing a dominant role in French, while vowel reduction and duration are more informative for intelligibility in Portuguese \cite{pinto2024acoustic}.

Automated multilingual dysarthria assessment studies reflect these language-dependent patterns. For example, feature importance analyses have shown that intelligibility prediction relies on different feature sets by languages, with prosodic features contributing more in English and segmental features playing a larger role in Korean \cite{yeo2022multilingual}. Multilingual models based solely on language-universal representations have been found to underperform monolingual systems \cite{kovac2022exploring, yeo2022cross}. Although incorporating both language-universal and language-specific features can outperform monolingual models \cite{yeo2022cross}, such approaches require additional training to support new languages, limiting their scalability.

\subsection{Universal Phone Recognizers}
Recent advances in Universal Phone Recognizers (UPRs) provide a language-agnostic foundation for phoneme-level speech analysis. UPRs generate phonetic transcriptions in the International Phonetic Alphabet (IPA), a standardized symbol system designed to represent speech sounds independently of any specific language. Trained on large-scale multilingual data, state-of-the-art UPRs have been shown to generalize across languages and to exhibit strong zero-shot performance on unseen languages and phonetic inventories \cite{zhu2025zipa, li2025powsm, Bharadwaj2026PRiSMBP}.

A key advantage of UPRs is that they provide direct access to segmental realizations of speech without relying on language models or decoding constraints, which typically impose lexical and phonotactic priors that can obscure pronunciation errors (\Cref{fig:fig1}). Prior work has shown that UPR-based transcriptions can capture consonant production deviations in English pediatric speech sound disorder, exhibiting moderate correlations ($\rho=0.63$) with SLP ratings of consonant correctness \cite{rosero2025finding}. However, this work was limited to English and focused on consonant-level transcription accuracy, and did not address language-specific interpretation of UPR outputs.


\begin{figure*}[t]
    \centering
    \includegraphics[width=0.65\paperwidth]{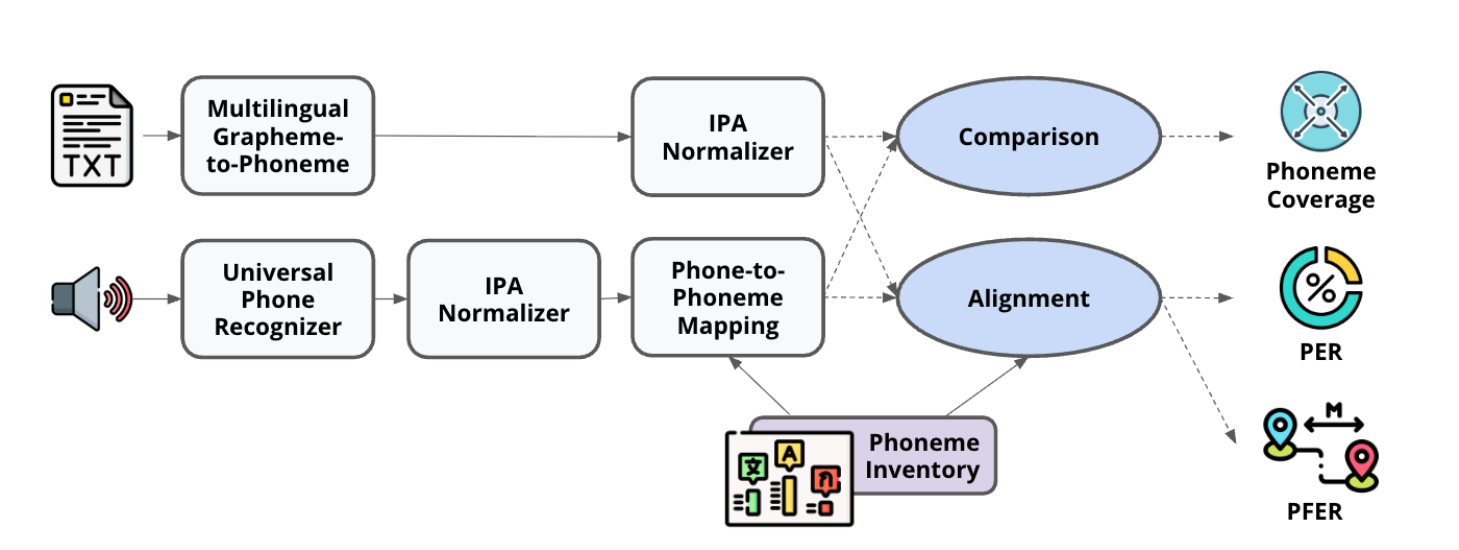}
    \caption{%
    \textbf{Overview of the framework.} Instantiation of the conceptual framework shown in Fig. 1 at the phoneme-production level.
(1) Universal phone recognition generates language-independent phone sequences. 
(2) Language-specific adaptation maps phones to phonemes, aligns predictions with references, and computes feature distances. 
(3) Three phoneme production metrics, including Phoneme Coverage (PhonCov), Phoneme Error Rate (PER), and Phonological Feature Error Rate (PFER), are computed and used to evaluate the framework against clinician-rated intelligibility scores.
    }
    \label{fig:summary}
\end{figure*}
\section{Method}\label{sec:method}
\subsection{Overview}
Our framework quantifies phoneme-level production accuracy through three key stages (\Cref{fig:summary}). First, a \textbf{universal phone recognizer} transcribes speech into language-agnostic phonetic representations. Second, \textbf{language-specific interpretation} maps these representations to each language’s phoneme inventory using contrastive phonological feature distances, supporting (i) phone-to-phoneme \textit{mapping} and, when applicable, (ii) contrastive phonological feature-aware \textit{alignment} between predicted and reference sequences. Finally, the framework \textbf{computes clinically interpretable metrics}, including Phoneme Error Rate (PER), Phonological Feature Error Rate (PFER), and Phoneme Coverage (PhonCov), which are evaluated against SLP-rated intelligibility scores. 
This pipeline operationalizes the two-stage conceptual framework shown in \Cref{fig:fig0}: the UPR corresponds to the universal speech model, while the contrast-aware mapping, alignment, and metric computation instantiate the language-specific intelligibility assessment module.

\subsection{Universal Phone Recognition and Multilingual G2P}
Speech is first transcribed into IPA using UPR, yielding language-agnostic phonetic representations. We employ two wav2vec2phoneme variants \cite{xu2021simple} and ZIPA \cite{zhu2025zipa}, which represent state-of-the-art approaches to universal phone recognition. Multiple UPR architectures are evaluated to assess the robustness of the proposed framework across recognizers. The specific model configurations are described below.


\begin{enumerate}
    \item \texttt{wav2vec2-lv60-espeak-cv-ft}\footnote{\url{https://huggingface.co/facebook/wav2vec2-lv-60-espeak-cv-ft}} is fine-tuned from the wav2vec2-large-LV60 model, which was trained on approximately 60{,}000 hours of English speech. Fine-tuning is conducted on Common Voice \cite{ardila2020common}, where orthographic transcriptions are converted into phoneme sequences using eSpeak\cite{espeak}. The model is fine-tuned with a Connectionist Temporal Classification (CTC) objective.

    \item \texttt{wav2vec2-xlsr-53-espeak-cv-ft}\footnote{\url{https://huggingface.co/facebook/wav2vec2-xlsr-53-espeak-cv-ft}} follows the same fine-tuning procedure as wav2vec2-lv60-espeak-cv-ft, but is initialized from the cross-lingually pretrained wav2vec2-XLSR-53, trained on unlabeled approximately 56{,}000 hours of speech from 53 languages.

    \item \texttt{ZIPA-large-crctc-800k}\footnote{\url{https://huggingface.co/anyspeech/zipa-large-crctc-ns-800k}} is a consistency-regularized CTC variant of ZIPA \cite{zhu2025zipa}, trained from scratch with a Zipformer-based encoder on 17{,}132 hours of phonetically transcribed speech spanning 88 languages \cite{zhu2024taste}, and further scaled using 11{,}851 hours of pseudo-labeled multilingual speech via noisy-student training.
\end{enumerate}

In parallel, reference phoneme sequences are generated via Epitran \cite{mortensen2018epitran}, multilingual grapheme-to-phoneme (G2P). Both UPR predictions and G2P outputs are normalized using \texttt{ipatok}\cite{ipatok} to ensure consistent tokenization of IPA symbols across languages (e.g., \textipa{\t{/ts/}} $\rightarrow$ \textipa{/ts/}). Diphthongs are treated as sequences of two monophthongs, to enable consistent comparison across languages with differing diphthong inventories.

\subsection{Language-Specific Phoneme Interpretation}\label{ssec:mapping}
Each language has its own phoneme inventory and corresponding phonemic contrasts. Consequently, identical phonetic realizations may be perceived differently depending on the listener’s native language \cite{kuhl1991human, kuhl2008phonetic, barrios2016establishing}, and thus have different impacts on intelligibility \cite{kim2017cross, yeo2023automatic}. Our framework models this phenomenon by performing language-specific interpretations of the language-universal IPA sequences generated by UPRs.


\subsubsection{Language-specific phonemic contrast identification}\label{sssec:contrastive-identification}
We first identify contrastive phonological features for each target language using its phoneme inventory and PanPhon \cite{Mortensen-et-al:2016}. PanPhon represents each IPA symbol as a vector of 24 articulatory features with values $+$ (present), $-$ (absent), or $0$ (not applicable), encoding features such as voicing, place, and manner of articulation. For instance, the unvoiced bilabial plosive [p] is represented as [$-$voice, $+$labial, $-$nasal, $+$anterior, $0$tense, $\cdots$]. A feature is considered contrastive in a given language if it assumes both $+$ and $-$ values within that language’s phoneme inventory.  For instance, Tamil exhibits a phonemic vowel length contrast; accordingly, the \texttt{long} feature is treated as contrastive in Tamil but not in languages lacking such a distinction.

\subsubsection{Distance calculation}\label{sssec:Distance}
Distances between segments are computed from normalized, weighted PanPhon representations using the phonemic contrasts identified in \Cref{sssec:contrastive-identification}. For each language, contrastive features are assigned full weight ($w_i = 1.0$), while non-contrastive features are ignored ($w_i = 0.0$), yielding a language-specific phonological feature distance that emphasizes phonemic structure. The distance between two segments $s_1$ and $s_2$ is defined as:
\[
d_{\text{feat}}(s_1, s_2; \mathbf{w}) =
\frac{\sum_{i=1}^{D} w_i \cdot |v_{s_1,i} - v_{s_2,i}|}
{\sum_{i=1}^{D} w_i \cdot 2},
\]
where $D = 24$ denotes the number of PanPhon features, $v_{s_1,i}$ and $v_{s_2,i}$ represent the $i$-th feature values for segments $s_1$ and $s_2$, and $\mathbf{w}$ is the vector of language-specific feature weights. The resulting distance is normalized to $[0,1]$, with 0 indicating identical segments and 1 indicating maximal featural divergence.
Distance is used in both the following \textit{phone-to-phoneme mapping} and \textit{sequence alignment}.

\subsubsection{Phone-to-phoneme mapping}
Native listeners tend to categorize speech sounds according to the phonemic categories of their native language, a phenomenon known as the perceptual magnet effect \cite{kuhl1991human, kuhl2008phonetic, barrios2016establishing}. Motivated by this principle, our framework maps the phonetic outputs of UPRs onto the phoneme inventory of the target language by assigning each predicted phone to the closest target-language phoneme under the weighted phonological feature distance defined in Section~\ref{sssec:Distance}.

In cases where multiple phonemes yield the same minimum distance, following tie-breaking rules are applied. Specifically,
(1) if the predicted phone contains diacritics, preference is given to phonemes matching the base form (e.g., \textipa{p\super h} $\rightarrow$ \textipa{p});
(2) phonemes with a greater number of matching contrastive features are preferred;
(3) for vowels, preference is given in the order of height, backness, and roundness, and for consonants, manner, place, and voicing are prioritized.\footnote{This ordering reflects established acoustic and perceptual findings regarding feature robustness and cross-linguistic variability. For vowels, height and backness correspond to the primary acoustic
dimensions of vowel quality, as reflected in F1-F2 structure, while rounding interacts with backness and modifies spectral structure \cite{stevens1998acoustic, johnson2012acoustic}.
For consonants, phoneme confusion studies show that place distinctions are particularly vulnerable under noise, whereas manner and voicing tend to be more robust depending on signal conditions \cite{miller1955analysis, wang1973confusions}. Additionally, realization of voicing contrasts varies across languages \cite{lisker1964vot}.}
and (4) remaining ties are resolved by selecting the phoneme with higher frequency in the target-language corpus. 


\subsubsection{Reference and prediction sequence alignment}
The alignment stage employs a weighted Needleman–Wunsch algorithm \cite{needleman1970general} to align reference G2P phoneme sequences with UPR-predicted sequences. Contrast-aware phonological feature distances defined in \Cref{sssec:Distance} are used as substitution costs, enabling the alignment to reflect phonological similarity rather than treating all segment substitutions uniformly. Under this formulation, segment pairs that differ on few contrastive features (e.g., $[\text{p}] \leftrightarrow [\text{b}]$) incur low substitution costs, whereas pairs that differ on many features (e.g., $[\text{p}] \leftrightarrow [\text{g}]$) incur higher costs. This feature-weighted design penalizes phonologically similar substitutions less than dissimilar ones, yielding linguistically informed alignments. In addition, to prevent spurious alignments between highly dissimilar segments, we introduce a distance threshold $\theta$: candidate substitutions with $d_{\text{feat}} > \theta$ are disallowed and instead realized as deletion–insertion events. Details of the selection of $\theta$ are provided in \Cref{sssec:selection_tau}.

\begin{table*}
\caption{Phoneme inventories and Contrastive features by language.}
\centering\label{tab:inventory}
\resizebox{0.9\textwidth}{!}{
\begin{tabular}{rr|l|l}
\toprule
Languages & Class &  Phoneme inventory & Contrastive features \\
\midrule
English & consonants & \textipa{/p, b, t, d, k, \textg, m, n, N, f, v, \texttheta, \dh, s, z, S, Z, h, \textteshlig, \textdyoghlig, l, \textturnr, j, w/} & [syl, son, cons, cont, delrel,
    lat, nas, strid, voi, ant, cor, distr, lab] \\
    & vowels &  \textipa{/i, I, e, E, \ae, A, O, o, U, u, 2, @/} & [hi, lo, back, round,
    tense] \\
    
Spanish &  consonants & \textipa{/p, b, t, d, k, g, m, n,                       \textltailn, f, \texttheta, s, x, J,
                    \textteshlig, l, \textturny, r, \textfishhookr/} & 
                    [syl, son, cons, cont, delrel, lat, nas, strid, voi, ant, cor, distr, lab] \\
 & vowels & \textipa{/i, e, a, o, u/} & [hi, lo, back, round] \\
        
Italian &  consonants & \textipa{/p, b, t, d, k, g, m, n, \textltailn, f, v, s, z, S, h, \texttslig, \textdyoghlig, \textteshlig, \textdyoghlig, l, \textturny, r, j, w/} & [syl, son, cons, cont, delrel, lat, nas, strid, voi, ant, cor, distr, lab]\\
        & vowels & \textipa{/i, e, E, a, O, o, u/} & [hi, lo, back, round, tense]  \\
        
Tamil   &  consonants &  
\textipa{/p, b, \textsubbridge{t}, \textsubbridge{d}, \textrtailt, \:d, k, g,  \textteshlig, \textdyoghlig,
    m, n, \textrtailn, 
    j, \textscriptv, r, \textfishhookr, l, \textrtaill/} & 
    [syl, son, cons, cont, delrel, lat, nas, strid, voi, ant, cor, distr, lab] \\
        & vowels & \textipa{/i, i\textlengthmark, e, e\textlengthmark, a, a\textlengthmark, o, o\textlengthmark, u, u\textlengthmark/} & 
        [hi, lo, back, round, long] \\
\bottomrule
\end{tabular}
}
\end{table*}


\subsection{Phoneme Production Metric Calculation}\label{ssec:metrics}
We compute three phoneme production metrics within the proposed framework: Phoneme Error Rate (PER), Phonological Feature Error Rate (PFER), and Phoneme Coverage (PhonCov). PER and PFER are derived from the alignment between reference and predicted sequences, whereas PhonCov is computed without alignment (\Cref{fig:summary}). All metrics are computed at the utterance level and subsequently aggregated at the speaker level.

\textbf{PER} measures overall phoneme production accuracy as the proportion of alignment errors relative to the length of the canonical phoneme sequence. It follows the standard definition of word error rate (WER) used in ASR evaluation, but is computed at the phoneme level rather than the word level. Given aligned reference and predicted sequences, PER is defined as
\[
\text{PER} = \frac{S + I + D}{N},
\]

\noindent where $S$, $I$, and $D$ denote the counts of substitutions, insertions, and deletions, respectively, and $N$ is the number of reference phonemes. PER ranges from 0 (perfect correspondence) to values exceeding 1 when insertions exceed the reference length. Lower values indicate more accurate phoneme production.

\textbf{PFER} quantifies the average phonological feature distance between aligned reference and predicted phonemes. It captures fine-grained deviations beyond the categorical errors reflected in PER. While PER treats every phoneme error as equal, PFER reflects how different the predicted phoneme is from its intended target in phonological feature space. For example, substituting /p/ with /b/ (differing in voicing only) results in a much smaller error score than substituting /p/ with /v/ (differing in voicing, place, and manner). 

After alignment between the canonical sequence $R = [r_1,...,r_N]$ and the predicted sequences $H = [h_1,...,h_N]$, PFER is defined as

\[
\text{PFER} = \frac{1}{N}\sum_{i=1}^{N} d_{\text{feat}}(r_i, h_i),
\]

\noindent where $r_i$ and $h_i$ denote the aligned reference and predicted segments, respectively, and $N$ is the length of the aligned sequence. $d_{\text{feat}}(r_i, h_i)$ corresponds to the distance defined in \Cref{sssec:Distance}. Deletions and insertions are assigned a distance of 1.0. Lower values indicate greater articulatory precision.

\textbf{PhonCov} quantifies phonemic diversity in speech production. Motivated by clinical evidence that speakers with dysarthria lose the ability to maintain phonemic contrasts as severity increases \cite{kim2011vowel, kim2010frequency}, PhonCov measures the extent to which a speaker’s productions realize the target language’s phoneme inventory. Specifically, it computes the proportion of reference phonemes that are realized at least once in the predicted material using an opportunity-normalized binary coverage formulation.

For each utterance, let $P_{\text{ref}}$ denote the set of unique phonemes appearing in the reference, and let $P^{hyp}$ denote the set of unique phonemes appearing in the corresponding predicted sequence. PhonCov is defined as:
\[
\text{PhonCov} (\%) = 100 \times \frac{|P_{\text{ref}}  \cap P_{\text{hyp}}|}{|P_{\text{ref}}|}.
\]

Unlike PER and PFER, PhonCov does not require sequence alignment, making it computationally efficient and less sensitive to alignment errors. In contrast to PER and PFER, higher PhonCov values indicates better phoneme production.


\subsection{Evaluation}
Speaker-level phoneme-production metrics are evaluated against clinical intelligibility scores using Kendall’s rank correlation coefficient ($\tau$). To assess whether language-specific processing yields significant changes in correlation strength, differences between dependent correlations are evaluated via bootstrap resampling over speakers (10{,}000 iterations). Statistical significance is determined by whether the 95\% confidence interval of the correlation difference excludes zero.


\section{Experiments}\label{sec:experiments}
\subsection{Datasets}
We analyzed four publicly available dysarthric speech datasets spanning English, Colombian Spanish, Italian, and Tamil. To eliminate the impact of prosody and syntactic predictability, we use only word-level materials.

\textbf{UASpeech}\cite{kim2008dysarthric} is an English corpus from 15 individuals with cerebral palsy and 13 age-matched speakers without impairments. The recorded word lists include 255 common words and 100 uncommon words. Each speaker is evaluated using speech intelligibility scores ranging from 1 (high intelligibility) to 4 (very low intelligibility), with healthy controls labeled as 0.

\textbf{PC-GITA}\cite{orozco2014new} is a Colombian Spanish speech dataset composed of 50 speakers with Parkinson’s disease and 50 healthy controls matched by age and gender. Participants produced 25 isolated words. Each speaker was evaluated using the UPDRS-speech scale, which rates speech impairment severity from 0 (normal) to 3 (severely affected). 

\textbf{Easycall}\cite{turrisi21_interspeech} is an Italian dataset including recordings from 26 speakers with dysarthria and 21 control speakers. The dysarthric group spans multiple etiologies, including Parkinson’s disease, Huntington’s disease, and ALS. Each participant produced 66 to 69 command words. Each speaker was analyzed with Theraphy Outcome Measurement, ranging from 1 (mild) to 5 (severe), with healthy speakers as 0.

\textbf{SSNCE}\cite{ta2016dysarthric} contains Tamil speech from 20 speakers with dysarthria, all with cerebral palsy, and 10 healthy speakers. The recordings consist of 103 unique words. Each speaker is rated for speech intelligibility, with scores ranging from 1 (high intelligibility) to 6 (low intelligibility), with healthy speakers rated as 0.

\subsection{Experiment Settings}
\begin{table*}[t!]

\caption{Kendall correlation ($\tau$) between metrics and clinical intelligibility scores. lv60 and xlsr53 refers to \texttt{wav2ph-lv60} and \texttt{wav2ph-xlsr53}, respectively. Avg. column is calculated across UPRs and languages.}
\label{tab:results}

\centering

\resizebox{\linewidth}{!}{

\begin{tabular}{l l  ccc|  ccc| ccc|  ccc | >{\columncolor{colorint!10}\arraybackslash}c}
\toprule
\multirow{2}{*}{\textbf{Metric}} &
\multirow{2}{*}{\textbf{Setting}} &
\multicolumn{3}{c|}{\textbf{English}} &
\multicolumn{3}{c|}{\textbf{Spanish}} &
\multicolumn{3}{c|}{\textbf{Italian}} &
\multicolumn{3}{c|}{\textbf{Tamil}} &
 \\


& & lv60 & xlsr53 & zipa
& lv60 & xlsr53 & zipa
& lv60 & xlsr53 & zipa
& lv60 & xlsr53 & zipa
& {\cellcolor{colorint!10}\textbf{Avg.}}  \\

\midrule


PER & raw 
& \textbf{0.820} & \textbf{0.820} & 0.828 & 0.263 & 0.406 & 0.068 & 0.636 & 0.643 & 0.681 & 0.752 & 0.783 & 0.732 & 0.619\\
 & mapping 
& \textbf{0.820} & \textbf{0.820} & \textbf{0.837} & \underline{\textbf{0.346}} & \textbf{0.449} & \underline{\textbf{0.323}} & \underline{0.666} & \underline{0.666} & 0.678 & \underline{0.813} & \underline{\textbf{0.833}} & 0.762 & 0.668 \\
 & alignment 
& \textbf{0.820} & \textbf{0.820} & 0.820 & 0.282 & 0.411 & \underline{0.123} & \underline{0.668} & \underline{0.671} & 0.694 & \underline{0.823} & 0.803 & \underline{0.767} & 0.642\\
 & mapping + alignment 
& \textbf{0.820} & \textbf{0.820} & 0.820 & \underline{\textbf{0.346}} & 0.419 & \underline{0.306}  & \underline{\textbf{0.686}} & \underline{\textbf{0.684}} & \underline{\textbf{0.704}} & \underline{\textbf{0.833}} & 0.828 & \underline{\textbf{0.788}} & \textbf{0.671}\\
\midrule

PFER & raw 
& 0.778 & \textbf{0.803} & 0.803 & 0.456 & 0.491 & 0.303 & 0.575 & 0.604 & 0.653 & 0.839 & 0.818 & 0.833 & 0.663\\
 & mapping 
& 0.778 & \textbf{0.803} & 0.803 & \textbf{0.457} & \textbf{0.493} & \underline{0.334} & 0.576 & 0.604 & 0.653 & 0.828 & 0.803 & 0.833 & 0.664\\
 & alignment
& 0.795 & 0.795 & \textbf{0.811} & 0.452 & 0.487 & \underline{0.389} & \underline{\textbf{0.636}} & \underline{\textbf{0.646}} & \underline{\textbf{0.701}} & 0.844 & 0.833 & \textbf{0.844} & \textbf{0.686}\\
 & mapping + alignment 
& \textbf{0.803} & 0.795 & \textbf{0.811} & 0.441 & 0.471 & \underline{\textbf{0.402}} & \underline{0.633} & \underline{0.643} & \underline{\textbf{0.701}} & \textbf{0.849} & \textbf{0.849} & 0.839 & \textbf{0.686}\\


\midrule

PhonCov & raw 
& \textbf{-0.786} & \textbf{-0.778} & \textbf{-0.795} & -0.295 & -0.421 & -0.187 & -0.431 & -0.581 & \textbf{-0.656} &  -0.747& -0.778 & -0.747 & -0.600\\
 & lang.-spec.\ mapping 
& \textbf{-0.786}  & \textbf{-0.778} & \textbf{-0.795} & \underline{\textbf{-0.407}} & \textbf{-0.472} & \underline{\textbf{-0.453}} & \textbf{-0.516} & \textbf{-0.591} & -0.623 & \textbf{-0.808} & \textbf{-0.818} & \textbf{-0.778} & \textbf{-0.652}\\

\bottomrule
\end{tabular}
}
\end{table*}

\subsubsection{Language-specific contrastive features}
We automatically identify language-specific contrastive features following the procedure described in \Cref{sssec:contrastive-identification}. As a result, for consonants, all four languages shared the contrastive features, while for vowels, English and Italian exhibit a tense–lax contrast (\texttt{tense}), while Tamil contrasts vowel length (\texttt{long}). These identified contrastive features are applied in distance calculations, including phone-phoneme mapping, sequence alignment, and PFER calculations. The phoneme inventories and corresponding contrastive features used in this study are summarized in \Cref{tab:inventory}. 

\subsubsection{$\theta$ selection}\label{sssec:selection_tau}
For each language, we compute contrastive-feature distances $d_\text{feat}(p_i, p_j)$ between all distinct consonant phonemes and define the nearest-neighbor distance of each consonant as the distance to its closest phonemic contrast\footnote{Vowel distances were consistently smaller than consonant distances across languages. To avoid forcing plausible consonant substitutions to be realized as deletion–insertion events, we derive $\theta$ from consonant distances.}. The substitution threshold $\theta$ is set to the 90th percentile of these nearest-neighbor distances, ensuring that substitutions between closely related consonants are permitted while highly dissimilar pairs are disallowed. The resulting language-specific thresholds used for alignment are: English (0.1667), Spanish (0.2118), Italian (0.1889), and Tamil (0.1944).

\subsection{Baseline experiments}
The goal of this work is not to optimize intelligibility prediction through supervised learning, but to evaluate training-free, interpretable phoneme-production metrics that can be applied to new languages without fitting a predictor. Accordingly, comparisons are restricted to widely used non-trained baselines.

The primary baseline is the three metrics (PER, PFER, and PhonCov) computed directly from the raw outputs of the UPRs without any language-specific processing, including phone-to-phoneme mapping or reference–prediction alignment. This setting provides a direct point of comparison for isolating the effects of language-specific phoneme interpretation. In the absence of language-specific alignment, PER and PFER are computed using Levenshtein distance \cite{levenshtein1966binary}  for sequence alignment, which assigns uniform costs to all substitutions, insertions, and deletions.

We further compare against commonly used acoustic feature–based baselines. Specifically, we extract the extended Geneva Minimalistic Acoustic Parameter Set (eGeMAPS) \cite{eyben2015geneva}, which comprises standardized prosodic, spectral, and voice quality features and has been widely used for intelligibility assessment \cite{van2023automatic, murton2023validation}. The resulting 88 acoustic features are each correlated individually with intelligibility scores, and we report both (i) the average correlation across features and (ii) the maximum correlation observed. In addition, we include cepstral peak prominence (CPP), an acoustic measure associated with impaired laryngeal control. Following \cite{murton2023validation}, CPP is computed over voiced segments only using Parselmouth \cite{jadoul2018introducing} with default settings.

As another baseline, we include ASR-based word error rate (WER), which has been reported to correlate with articulatory imprecision and reduced intelligibility in dysarthric speech \cite{XUE202323}. We employ multilingual Whisper-small and Whisper-medium models, selected to be comparable in scale (approximately 300M parameters) to the UPRs used in this study.


\section{Results}\label{sec:results}
\subsection{Correlations with Intelligibility Scores}\label{ssec:correlation}
Table \ref{tab:results} reports Kendall’s $\tau$ correlations between phoneme production metrics and clinical intelligibility scores. Bold values indicate the best-performing setting within each metric–language combination, while underlining marks improvements over the raw baseline whose confidence intervals exclude zero (\textit{i.e.} statistically significant). 

Across metrics and languages, incorporating language-specific processing generally improves correlations with intelligibility, though the magnitude and source of gains vary by metric. For PER, the combination of language-specific mapping and alignment yields the highest average performance across languages, with consistent improvements over the raw baseline in Spanish, Italian, and Tamil. 

For PFER, improvements are driven primarily by alignment rather than mapping. Alignment-based settings consistently outperform the raw baseline, particularly in Italian and Tamil, and achieve the highest average correlations across languages. In contrast, mapping alone yields smaller and less consistent gains, suggesting that PFER benefits mainly from explicit temporal alignment informed by phonological feature distances.

For PhonCov, language-specific mapping improves correlations over the raw setting. Notably, despite being alignment-free, PhonCov with mapping achieves correlations comparable to those of alignment-based metrics, highlighting its effectiveness as a lightweight measure of phonemic production.

English exhibits little to no improvement from language-specific processing. This observation is consistent with prior findings that UPR models trained using G2P-derived transcriptions perform phoneme recognition rather than phone recognition for sociophonetic variants of English \cite{zhu2025zipa}. Consequently, raw UPR outputs in English are already close to phoneme-level representations, leaving limited room for further gains from language-specific phoneme interpretation.

\begin{table}[t]

\caption{Absolute Kendall's $\tau$ between baseline features and intelligibility scores. Best performance per language is highlighted in bold.}

\centering

\resizebox{\linewidth}{!}{

\begin{tabular}{lcccc}

\toprule
\textbf{Features} & \textbf{English} & \textbf{Spanish} & \textbf{Italian} & \textbf{Tamil} \\
\midrule
CPP & 0.188 & 0.216 & 0.236 & 0.168 \\
eGeMAPS (avg.) & 0.271 & 0.146 & 0.253 & 0.270 \\
eGeMAPS (max.) & 0.709 & \textbf{0.364} & 0.585 & 0.544 \\
Whisper-small & \textbf{0.768} & 0.157 & 0.594 & \textbf{0.576} \\
Whisper-medium & 0.755 & 0.345 & \textbf{0.638} & 0.552 \\
\bottomrule
\end{tabular}\label{tab:baseline}
}
\end{table}

\subsection{Comparison to Baseline Approaches}\label{ssec:baseline}
Across languages, our UPR-based phoneme-production metrics (\Cref{tab:results}) consistently achieve higher correlations with clinical intelligibility scores than all baseline features (\Cref{tab:baseline}). Absolute values are reported because eGeMAPS features can show either positive or negative associations with severity. This indicates that phoneme-level measures capture intelligibility degradation more effectively than generic acoustic or word-level features.

Among the baseline acoustic features, CPP exhibits uniformly weak correlations with intelligibility across languages. This is expected, as CPP captures intelligibility-related information in a highly restricted manner, focusing primarily on laryngeal function. Similarly, eGeMAPS features yield low correlations when averaged across the full feature set. In contrast, the strongest individual eGeMAPS features achieve substantially higher correlations, suggesting that intelligibility is driven by a limited subset of acoustic cues. This observation is consistent with prior work showing that feature selection over eGeMAPS improves correlations with intelligibility scores \cite{van2023automatic}.
ASR-based WER exhibits stronger correlations with intelligibility than acoustic baselines in English, Italian, and Tamil,  potentially reflecting that articulation is among the strongest contributors to intelligibility degradation in pathological speech \cite{xue2023assessing}. Nevertheless, WER remains less predictive than the proposed phoneme-production metrics.

\section{Analysis}\label{sec:analysis}
\subsection{Comparisons across Universal Phone Recognizers}
No single UPR architecture consistently outperforms the others across languages. In English, correlations are near ceiling across all recognizers and processing settings, resulting in negligible inter-model differences. Spanish exhibits substantial UPR-dependent variability under raw settings, which is reduced by language-specific mapping and alignment. Italian shows smaller but noticeable inter-recognizer differences that are further stabilized by language-specific processing, whereas Tamil is already relatively stable across UPRs, with language-specific interpretation yielding mainly modest gains in average performance rather than pronounced reductions in variability. Overall, these results indicate that language-specific phoneme interpretation improves the stability of intelligibility prediction and reduces sensitivity to the underlying UPR architecture, particularly in languages with higher baseline variability.

\subsection{Impact of Non-contrastive Features}
To assess the contribution of non-contrastive phonological features to intelligibility prediction, we vary the weight assigned to non-contrastive features in the phonological feature distance calculation ($w_{nc} \in \{0, 0.5, 1\}$, where nc denotes non-contrastive). A value of $w_{nc} = 0$ restricts distance computation to contrastive features only, whereas $w_{nc} = 1$ assigns equal weight to contrastive and non-contrastive features. We evaluate the effect of this weighting across metrics and processing settings, including mapping, alignment, and their combination.

PER and PhonCov are unaffected by variations in \( w_{nc} \) across languages and settings. In contrast, PFER exhibits modest sensitivity to \( w_{nc} \), where increasing \( w_{nc} \) generally leads to small reductions in correlation with clinical intelligibility scores. This effect is most pronounced in Italian, which shows consistent reductions of approximately 2–3\% across processing modes, and in Spanish, which exhibits a larger reduction of approximately 5\% in the alignment-only setting. English and Tamil remain largely unaffected, with changes below 2\% or negligible across modes.

Overall, these results indicate that non-contrastive features contribute little to intelligibility prediction within this framework and, when influential, tend to reduce rather than improve performance. The invariance of PER and PhonCov further suggests that discrete, binary-based metrics are robust to the inclusion of non-contrastive features, whereas distance-based metrics such as PFER are more sensitive to their weighting. These findings support the use of contrastive features alone, enabling simpler feature representations without loss of predictive accuracy. Because the languages examined here share broadly similar contrastive feature inventories, further evaluation on typologically more diverse languages is required to fully assess the role of non-contrastive features.

\subsection{Interactive Contributions to Intelligibility Severity}\label{ssec:regression}
We analyzed the joint contributions of UPR-based phoneme-production metrics to intelligibility using nested ordinal logistic regression, using the best-performing configuration for each metric: PER with mapping+alignment, PFER with alignment only, and PhonCov with mapping. 

In English, the PER-only model achieved a high pseudo-$R^2$ (0.785, $p < 0.001$), and adding PFER with its interaction term produced a near-ceiling fit ($\Delta R^2 = 0.215$, $p = 0.0006$), while PhonCov did not provide additional improvement.
In contrast, non-English languages show consistent complementary contributions from all three metrics. In Spanish, the PER-only model achieved a pseudo-$R^2$ of 0.153 ($p < 0.001$), and adding PFER improved model fit ($\Delta R^2 = 0.073$, $p < 0.001$), with the full model including PhonCov achieving the best performance ($\Delta R^2 = 0.046$, $p = 0.0412$). Italian exhibited a similar pattern, with the PER-only model achieving a pseudo-$R^2$ of 0.383 ($p < 0.001$), a modest gain from PFER ($\Delta R^2 = 0.040$, $p = 0.0498$) followed by an improvement from PhonCov ($\Delta R^2 = 0.119$, $p = 0.0014$). In Tamil, the PER-only model achieved a pseudo-$R^2$ of 0.618 ($p < 0.001$), and adding PFER provided marginal improvement ($\Delta R^2 = 0.057$, $p = 0.0512$). Incorporating PhonCov yielded an additional gain ($\Delta R^2 = 0.188$, $p < 0.001$).

\subsection{Phoneme-level Analysis}
\Cref{ssec:regression} demonstrates that PER, PFER, and PhonCov provide complementary contributions to intelligibility prediction. We further examine the types of information captured by each metric and assess their correspondence with clinical observations of dysarthria reported in prior studies. All analyses use ZIPA outputs and the same best-performing configurations used in \Cref{ssec:regression}.


\begin{figure}[t]
    \centering
    \includegraphics[width=0.85\columnwidth]{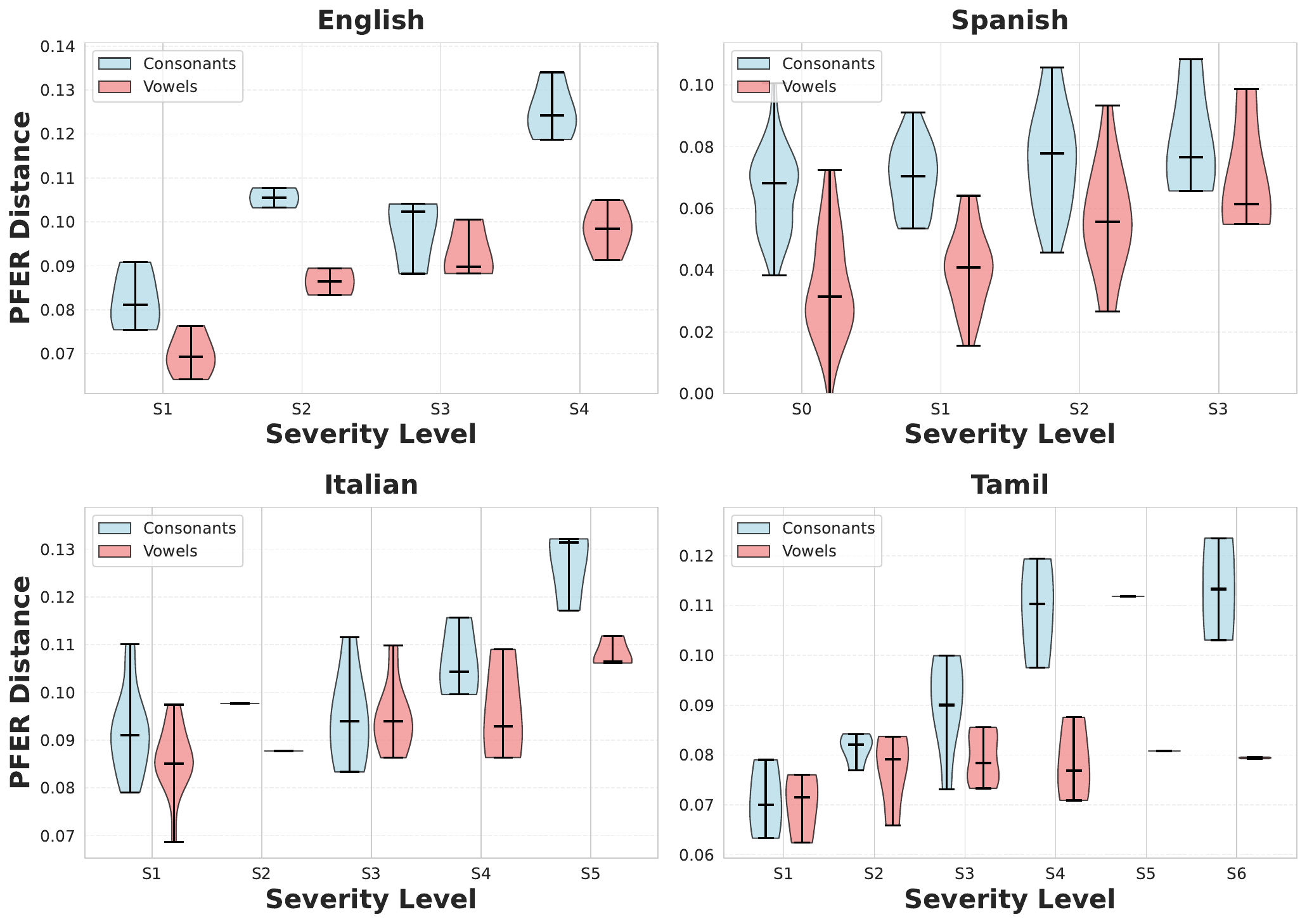}
    \caption{Distribution of phoneme feature error distances of consonants across severity levels. Violin plots show the density of PFER, with overlaid mean and median values. S denotes severity, and the following number indicates the severity level.
    }  
    \label{fig:pfer}
\end{figure}

\subsubsection{PER analysis}
While PER summarizes overall disruption in phoneme production as a single value, the error types that constitute its calculation, including correct matches, substitutions, deletions, and insertions, provide additional insight into how phoneme-level production errors vary with dysarthria severity. To quantify these patterns, we aggregate PER error-type counts by severity level and normalize them by the total number of reference phonemes.

Across languages, error distributions change with increasing severity. The proportion of correct phoneme productions decreases from approximately 62–72\% at the lowest severity levels to 13–58\% at the highest levels. Deletions show the strongest association with severity, increasing from roughly 6–14\% to 20–61\% and becoming the dominant error type at higher severity levels. This pattern is consistent with clinical reports that severe dysarthria is characterized by increased phoneme omissions \cite{kim2010frequency}.

In contrast, substitution rates remain relatively stable across severity levels, ranging from 18–32\%, suggesting that phoneme confusions occur throughout the severity spectrum rather than increasing monotonically with impairment. Insertion rates vary substantially across languages, ranging from 6–40\%. These insertions may reflect language-specific factors such as syllable structure \cite{levelt2004syllable}, though further investigation is required to clarify their role in dysarthric speech.

\subsubsection{PFER analysis}
PFER represents errors as graded phonological feature distances and encodes the magnitude of deviation between produced and target phonemes. Clinically, dysarthria severity is associated with progressively larger articulatory deviations \cite{kim2010frequency}, in addition to increased error frequency. To examine whether PFER reflects this graded deterioration, we analyzed speaker-level PFER distributions across severity levels (\Cref{fig:pfer}).

Across languages, PFER distributions increase and become more dispersed with dysarthria severity, indicating progressively larger and more variable phonological deviations as intelligibility declines. Consonant PFER values are consistently higher and increase more sharply than vowel PFER, which remain lower and change more gradually. This pattern suggests that consonant production undergoes greater degradation with increasing severity, consistent with the clinical emphasis on consonant accuracy in intelligibility assessment, where percentage of consonants correct (PCC) is commonly used as a proxy for intelligibility \cite{shriberg1997percentage, sell2020percent}.

\begin{table}[t]
\caption{Top two phonemes with highest and lowest discriminability across intelligibility severity levels. S0 and Smax denote healthy and the most severe, respectively.}
\centering
\resizebox{0.9\linewidth}{!}{%
\label{tab:phoneme_discriminability}
\begin{tabular}{lllc c c}
\toprule
Language & Category & Rank & Phoneme & Slope & Covered (S0$\rightarrow$Smax) \\
\midrule
English & Top    & 1 & \textipa{Z} & -23.08 & 90.4\% $\rightarrow$ 4.4\% \\
& Top    & 2 & \textipa{z} & -22.95 & 89.5\% $\rightarrow$ 4.9\% \\
& Lowest & 1 & \textipa{A} & -6.58  & 88.5\% $\rightarrow$ 55.6\% \\
 & Lowest & 2 & \textipa{i} & -11.60 & 94.1\% $\rightarrow$ 41.3\% \\
\midrule
Spanish & Top    & 1 & \textipa{k} & -6.80  & 88.8\% $\rightarrow$ 55.0\% \\
 & Top    & 2 & \textipa{e} & -5.39  & 86.0\% $\rightarrow$ 55.0\% \\
 & Lowest & 1 & \textipa{u} & -0.27  & 85.7\% $\rightarrow$ 66.7\% \\
 & Lowest & 2 & \textipa{m} & -0.83  & 98.4\% $\rightarrow$ 81.2\% \\
\midrule
Italian & Top    & 1 & \textipa{v} & -14.81 & 86.0\% $\rightarrow$ 12.0\% \\
 & Top    & 2 & \textipa{r} & -13.21 & 84.9\% $\rightarrow$ 19.6\% \\
 & Lowest & 1 & \textipa{a} & -3.08  & 91.4\% $\rightarrow$ 72.5\% \\
& Lowest & 2 & \textipa{n} & -7.97  & 85.7\% $\rightarrow$ 44.1\% \\
\midrule
Tamil   & Top    & 1 & \textipa{u:} & -14.15 & 90.0\% $\rightarrow$ 0.0\% \\
  & Top    & 2 & \textipa{l}  & -12.68 & 97.0\% $\rightarrow$ 5.0\% \\
   & Lowest & 1 & \textipa{i}  & -5.90  & 83.3\% $\rightarrow$ 37.5\% \\
   & Lowest & 2 & \textipa{e}  & -9.10  & 80.0\% $\rightarrow$ 20.0\% \\
\bottomrule
\end{tabular}
}
\vspace{-5mm}
\end{table}

\subsubsection{PhonCov analysis}
To identify phonemes most sensitive to intelligibility degradation, we analyzed phoneme-level coverage across severity levels. Sensitivity was quantified using the absolute slope obtained from a weighted linear regression of coverage on severity, where weights corresponded to the number of speakers at each severity level to account for varying sample sizes. Larger absolute slope values indicate stronger severity-dependent change. Only phonemes with coverage $\geq$ 80\% in healthy speech were included to minimize errors introduced by UPR predictions (\Cref{tab:phoneme_discriminability}).

Although phoneme-specific patterns varied across languages, consistent trends emerged. Phonemes requiring greater articulatory complexity exhibited the highest discriminability, including voiced fricatives (\textipa{Z, z, v}) and liquids (\textipa{r, l}). According to the four-level articulatory complexity framework proposed by \cite{kent1992biology}, these phonemes correspond to complexity levels 3 and 4, which involve increased articulatory coordination and fine temporal control. In contrast, phonemes with the lowest discriminability, such as corner vowels (\textipa{A, a, u, i, e}) and bilabial nasals, align with complexity level 1, characterized by relatively simple and stable articulatory gestures. Notably, in Spanish, the slopes of the top two phonemes fall within the range of the lowest discriminability observed in other languages, suggesting reduced overall sensitivity of phoneme-level coverage to severity in this language.

Taken together, the results indicate that phonemes with greater articulatory complexity are more susceptible to dysarthric degradation, whereas simpler phonemes remain relatively stable across severity levels. This suggests that articulatory complexity may be a useful criterion for weighting phonemes in intelligibility prediction models and for prioritizing targets in intelligibility monitoring protocols.

\section{Discussion}\label{sec:discussion}
The proposed UPR-based phoneme-production metrics capture clinically relevant and complementary dimensions of phoneme production in dysarthric speech. Each metric targets a distinct aspect of segmental impairment and is therefore suited to different analytical and clinical objectives. 

\textbf{PER} quantifies the frequency of phoneme-level errors and serves as a robust indicator of overall segmental breakdown, making it well suited as a global proxy for intelligibility severity. \textbf{PFER}, in contrast, models phoneme errors as graded distances in phonological feature space, capturing the magnitude of articulatory deviation and providing sensitivity to subtle phonetic distortions that are not reflected in discrete error counts. \textbf{PhonCov} measures the breadth of phoneme realization independent of alignment, offering a low-complexity yet clinically interpretable summary of phonemic coverage. 

Importantly, given language-specific deviations, these metrics are best interpreted as \emph{relative measures} that characterize how phoneme production degrades with increasing severity rather than as absolute indices. This property makes them well suited for severity stratification, within-language cross-speaker comparison, and longitudinal monitoring of disease progression or treatment response.

\section{Conclusion}\label{sec:conclusion}
This study provides evidence that UPR-based phoneme-production metrics constitute a viable and interpretable approach to intelligibility assessment in dysarthric speech, exhibiting significant correlations with clinical intelligibility scores. Incorporating language-specific phoneme interpretation improves the relevance of these metrics by aligning universal phone outputs with the phonemic structure of the target language, yielding stronger associations with intelligibility than raw UPR outputs. 

Our results further show that the impact of language-specific processing depends on the metric. Discrete metrics benefit primarily from phone-to-phoneme mapping, consistent with perceptual categorization effects whereby listeners interpret speech sounds through native-language phonemic categories \cite{kuhl1991human, kuhl2008phonetic, barrios2016establishing}. In contrast, distance-based metrics such as PFER benefit most strongly from phonologically informed alignment, reflecting their sensitivity to minimal deviations such as phonetic distortions. Finally, PhonCov is notable in that it achieves performance comparable to alignment-based metrics despite its simplicity and lack of explicit alignment, highlighting the value of clinically motivated, low-complexity metrics.

Several limitations suggest directions for future work. First, the proposed framework focuses exclusively on segmental production and does not account for suprasegmental factors such as rhythm, timing, and intonation. Incorporating prosodic modeling may provide a more comprehensive account of intelligibility degradation and improve multilingual intelligibility assessment. Second, each language in this study is represented by a single dataset. Further validation on additional corpora is therefore necessary to assess generalizability and disentangle language effects from dataset-specific characteristics.

\section*{ACKNOWLEDGMENT}
The authors thank Dr.\ David Harwath and Dr.\ Jun Wang for their valuable discussions. This work was supported by the Texas Health Catalyst, Dell Medical School, UT Austin.


\balance

\bibliographystyle{IEEEtran}
\bibliography{mybib}





\balance

\end{document}